\pdfoutput=1

\documentclass[11pt]{article}

\usepackage[final]{emnlp2021}

\usepackage{times}
\usepackage{latexsym}
\usepackage{graphicx}
\usepackage{hyperref}
\usepackage{url}
\usepackage{booktabs}
\usepackage{graphicx}
\usepackage{wrapfig}
\usepackage{xcolor}
\usepackage{latexsym}
\usepackage{graphicx}
\usepackage{tabularx}
\usepackage{multirow}
\usepackage{booktabs}
\usepackage{xcolor}
\usepackage{adjustbox}
\usepackage{fancybox}
\usepackage{ulem}
\usepackage{wrapfig,lipsum,booktabs}
\usepackage{amsmath}
\usepackage{array}
\usepackage{makecell}
\usepackage{amsfonts}
\usepackage{bm}
\usepackage{amsmath}
\usepackage{amssymb}
\usepackage{subcaption}
\usepackage{footnote}
\usepackage{soul}

\usepackage[T1]{fontenc}

\usepackage[utf8]{inputenc}

\usepackage{microtype}

\definecolor{barca-blue}{RGB}{0, 76, 153}
\definecolor{barca-red}{RGB}{167, 0, 66}
\definecolor{lblue}{RGB}{13, 152, 255}
\definecolor{lred}{RGB}{255, 108, 108}
\definecolor{dblue}{RGB}{0, 112, 192}
\definecolor{dred}{RGB}{192, 0, 0}
\definecolor{Gray}{gray}{0.9}
\definecolor{tblue}{HTML}{174992}

\newcommand{\authorInfo}[1]{
\begingroup
\let\thefootnote\relax
\footnotetext{#1}
\endgroup}

\title{STT: Soft Template Tuning for Few-Shot Adaptation}

\author{
  Ping Yu \textsuperscript{*}\\
  University at Buffalo\\
  \texttt{pingyu@buffalo.edu} \\\And
  Wei Wang \textsuperscript{*}\\
  University at Buffalo\\
  \texttt{wwang49@buffalo.edu} \\\And
  Chunyuan Li \\
  Microsoft Research \\
  \texttt{lichunyuan24@gmail.com} \\\AND
  Ruiyi Zhang \\
  Adobe Research\\
  \texttt{ryzhang.cs@gmail.com} \\\And
  Zhanpeng Jin \\
  University at Buffalo\\
  \texttt{zjin@buffalo.edu} \\\And
  Changyou Chen \\
  University at Buffalo\\
  \texttt{changyou@buffalo.edu}
}

\begin{document}
\maketitle
\authorInfo{* These authors contributed equally to this work.}

\begin{abstract}
Prompt tuning has been an extremely effective tool to adapt a pre-trained model to downstream tasks. 
However, standard prompt-based methods mainly consider the case of sufficient data of downstream tasks. It is still unclear whether the advantage can be transferred to the few-shot regime, where only limited data are available for each downstream task. Although some works have demonstrated the potential of prompt-tuning under the few-shot setting, the main stream methods via searching discrete prompts or tuning soft prompts with limited data are still very challenging. Through extensive empirical studies, we find that there is still a gap between prompt tuning and fully fine-tuning for few-shot learning. To bridge the gap, we propose a new prompt-tuning framework, called Soft Template Tuning (STT). STT combines manual and auto prompts, and treats downstream classification tasks as a masked language modeling task. Comprehensive evaluation on different settings suggests STT can close the gap between fine-tuning and prompt-based methods without introducing additional parameters. Significantly, it can even outperform the time- and resource-consuming fine-tuning method on sentiment classification tasks. 
\end{abstract}

\section{Introduction}

With the success of pre-trained large language models, an increasing number of techniques have been proposed to adapt these general-propose models to downstream tasks. Starting from GPT \citep{radford2018improving} and BERT \citep{devlin2018bert}, \textbf{full model fine-tuning} is used as the default method to adapt pre-trained language models to downstream tasks. With the rapid increase of model sizes \citep{lewis2019bart,liu2019roberta,raffel2019exploring}, it has gradually become more challenging to update all model parameters during fine-tuning. 

One extreme case is the GPT-3 model~\citep{gpt3}, where the model size is too large (175B model parameters) to even enable fine-tuning, which sets a barrier for the community to involve in the research. Alternatively, \textbf{in-context learning} is proposed, which demonstrates few-shot capabilities without tuning any model parameters. However, although in-context learning enables jumbo language models to be applied on diverse downstream tasks, it is not as effective as fine-tuning, which significantly restricts the advantages and applicability of these large pre-trained language models.

Meanwhile, ideas have been explored to update a small number of model parameters while keeping most parameters frozen. \citet{stanford} proposed \textbf{prefix-tuning} which shows strong performance on generative tasks. This method freezes the model parameters and propagates the error during fine-tuning to prefix activation prepended to each layer in the encoder stack, including the input layer. \citet{hambardzumyan2021warp} simplified this method by restricting the trainable parameters only to the input and output subnetworks of a masked language model, and showed reasonable results on classification tasks. \citet{google} proposed a further simplification -- \textbf{prompt-tuning} for adapting language models, which shows that prompt tuning alone (with no intermediate-layer prefixes) is competitive with model tuning. However, \citet{google} only test prompt tuning in the case where each task has enough data for adaptation. 



\vspace{-10pt}
\paragraph{Why Prompt Tuning in the Few-Shot Regime}
To the best of our knowledge, only tuning prompt tokens for few-shot adaptation is under-explored. Note that related works have different settings with ours. For example, \citet{liu2021gpt} and \citet{han2021ptr} combines both discrete prompts and soft prompts, however, they fine-tune the soft prompts jointly with the entire pre-trained model; P-tuning v2 \cite{liu2021p} further enhanced the universality of \citet{liu2021gpt} on a broader selection of models and tasks at the cost of introducing soft prompts (more trainable parameters) to every layer of the pre-trained model, and is therefore only evaluated in the fully-supervised setting rather than few-shot setting. \citet{gu2021ppt} proposed Pre-trained Prompt Tuning framework (PPT). They proved that after pre-training on soft prompts, prompt tuning could reach or even outperform full-model fine-tuning under few-shot settings. However, the improvement was achieved at the cost of introducing additional pre-training efforts, 
moreover different types of pre-training tasks need to be recruited to accommodate diverse downstream tasks (e.g., single sentence, sentence-pair, multiple-choice classifications). For a comprehensive review of popular prompt-tuning based methods, please refer to Section~\ref{sec:related_works} in the Appendix. 
We find through empirical evidence that prompt tuning lags far behind the fine-tuning in the few-shot regime. This can be explained by the intuition that using only a few samples to learn a relatively small portion of parameters might be too flexible and tend to over-fit the few training data, similar to the traditional machine-learning setting. 

\begin{figure*}
    \centering
    \includegraphics[width=.9\linewidth]{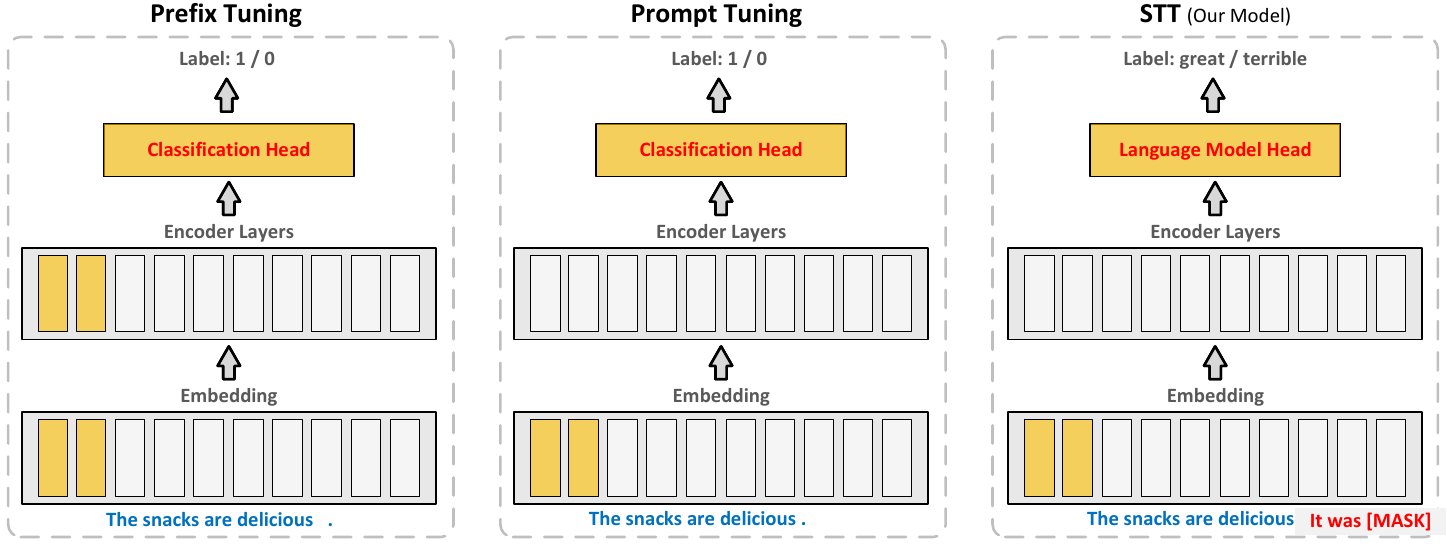}
    \caption{\textbf{Model Comparison:} We compare STT with Prefix tuning \citep{stanford} and Prompt tuning \citep{google}. The yellow box means parameters need to be updated. The white box means parameters are fixed. STT, on the basis of the Prompt tuning method, adds a human-defined template to the input sentence, and replaces the classifier head with the language model head to predict whether the word in the mask position is great or terrible.}
    \vspace{-3mm}
    \label{fig:model_structure}
\end{figure*}

\vspace{-10pt}
\paragraph{Our Proposed Method}
We propose improved techniques to alleviate this problem in the practical scenario where a moderate-sized pre-trained language model ({\it e.g.}, RoBERTa large \citep{roberta}) and a small number of examples for each downstream task ({\it i.e.}, the \textit{few-shot} setting) are accessible. This setting is common in practice: (1) Such types of models are publicly available and the computational resources needed are easily accessible for most researchers; (2) The few-shot settings are realistic, as new tasks usually come with few examples. To address the problems mentioned above, we explore the idea of closing the gap between pre-training and fine-tuning. Specifically, we propose to add manual prompts on the basis of prompt-tuning \cite{google} and treat the problem to be a masked language modeling problem, as was done by most pre-training. Since we combine manual template with soft prompt, we call our method Soft Template Tuning (STT). Experiments demonstrate that STT is able to close the gap between prompt-based methods and fine-tuning in the few-shot regime. Remarkably, STT can even outperform the computationally heavy fine-tuning on some tasks, though it tunes much fewer parameters than the fully fine-tuning method.

\begin{table*}[ht!]
    \centering
    \scalebox{0.7}{
    \begin{tabular}{l c|c c c c c c c c c |c}
    \toprule[1.2pt]
        & \textbf{Models} & \makecell{\textbf{SST-2}\\{\footnotesize (acc)}} & \makecell{\textbf{SST-5}\\{\footnotesize (acc)}} & \makecell{\textbf{MR}\\{\footnotesize (acc)}}& \makecell{\textbf{CR}\\{\footnotesize (acc)}} & \makecell{\textbf{MPQA}\\{\footnotesize (acc)}} & \makecell{\textbf{TREC}\\{\footnotesize (acc)}}& \makecell{\textbf{SNLI}\\{\footnotesize (acc)}}& \makecell{\textbf{QNLI}\\{\footnotesize (acc)}}& \makecell{\textbf{QQP}\\{\footnotesize (F1)}} & \textbf{Average}\\
    \midrule[0.8pt]
        \multirow{5}{*}{\rotatebox[origin=c]{90}{\textbf{One-shot}}}
        & Fine-tuning & 
        53.1\footnotesize{$\pm$4.4} & 23.4\footnotesize{$\pm$1.8} & 53.3\footnotesize{$\pm$3.6} & 52.8\footnotesize{$\pm$2.8} & 51.3\footnotesize{$\pm$1.8} & 28.7\footnotesize{$\pm$10.6} & 35.2\footnotesize{$\pm$2.6} & 51.0\footnotesize{$\pm$1.0} & 45.4\footnotesize{$\pm$11.4} & 43.8\\
        \cmidrule(lr){2-12}
        & Adapter & 51.0 \footnotesize{$\pm$2.9} & 18.9\footnotesize{$\pm$5.9} & 50.2\footnotesize{$\pm$0.5} & 49.7\footnotesize{$\pm$0.8} & 50.0\footnotesize{$\pm$0.1} & 23.1\footnotesize{$\pm$4.4} & 33.3\footnotesize{$\pm$0.3} & 50.3\footnotesize{$\pm$1.9} & 40.2\footnotesize{$\pm$7.9} & 39.5\\
        \cmidrule(lr){2-12}
        & Prefix-tuning & 
        51.0\footnotesize{$\pm$2.8} & 22.9\footnotesize{$\pm$3.9} & 51.2\footnotesize{$\pm$2.2} & 51.0\footnotesize{$\pm$2.7} & 53.0\footnotesize{$\pm$1.9} & \textcolor{tblue}{\textbf{27.9}}\footnotesize{$\pm$6.1} & 33.5\footnotesize{$\pm$0.8} & 49.8\footnotesize{$\pm$1.6} & 46.4\footnotesize{$\pm$4.9} &  42.9 \\
        & Prompt-tuning & 
        51.5\footnotesize{$\pm$2.2} & 22.7\footnotesize{$\pm$2.8} & 51.5\footnotesize{$\pm$2.1} & 52.1\footnotesize{$\pm$3.5} & 51.9\footnotesize{$\pm$2.1} & 17.3\footnotesize{$\pm$3.3} & \textcolor{tblue}{\textbf{35.1}}\footnotesize{$\pm$2.0} & 50.6\footnotesize{$\pm$0.5} & \textcolor{tblue}{\textbf{47.8}}\footnotesize{$\pm$8.7} & 42.3 \\
        & STT (ours) & 
        \textcolor{tblue}{\textbf{57.3}}\footnotesize{$\pm$5.6} & \textcolor{tblue}{\textbf{23.5}}\footnotesize{$\pm$3.3} & \textcolor{tblue}{\textbf{52.1}}\footnotesize{$\pm$1.3} & \textcolor{tblue}{\textbf{55.5}}\footnotesize{$\pm$5.1} & \textcolor{tblue}{\textbf{55.2}}\footnotesize{$\pm$3.7} & 26.0\footnotesize{$\pm$11.7} & 33.8\footnotesize{$\pm$1.5} & \textcolor{tblue}{\textbf{51.8}}\footnotesize{$\pm$1.0} & 45.4\footnotesize{$\pm$11.9} & \textcolor{tblue}{\textbf{44.5}}\\
    \midrule[0.8pt]
        \multirow{5}{*}{\rotatebox[origin=c]{90}{\textbf{Two-shot}}}
        & Fine-tuning &  
        55.2\footnotesize{$\pm$8.8} & 27.0\footnotesize{$\pm$2.8} & 55.9\footnotesize{$\pm$3.0} & 58.5\footnotesize{$\pm$3.4} & 55.6\footnotesize{$\pm$6.3} & 43.8\footnotesize{$\pm$6.9} & 36.4\footnotesize{$\pm$3.5} & 52.3\footnotesize{$\pm$1.2} & 53.4\footnotesize{$\pm$4.6} & 48.7\\
        \cmidrule(lr){2-12}
        & Adapter & 50.2 \footnotesize{$\pm$0.8} & 17.9\footnotesize{$\pm$6.5} & 50.0\footnotesize{$\pm$0.0} & 50.0\footnotesize{$\pm$0.0} & 50.0\footnotesize{$\pm$0.1} & 20.0\footnotesize{$\pm$7.3} & 33.4\footnotesize{$\pm$0.4} & 50.4\footnotesize{$\pm$0.5} & 40.2\footnotesize{$\pm$11.8} & 40.2\\
        \cmidrule(lr){2-12}
        & Prefix-tuning &  
        51.7\footnotesize{$\pm$2.4} & 23.5\footnotesize{$\pm$2.3} & 52.1\footnotesize{$\pm$1.8} & 53.4\footnotesize{$\pm$3.6} & 55.3\footnotesize{$\pm$4.4} & \textcolor{tblue}{\textbf{35.8}}\footnotesize{$\pm$4.0} & \textcolor{tblue}{\textbf{34.4}}\footnotesize{$\pm$0.6} &  51.3\footnotesize{$\pm$2.5} & 46.0\footnotesize{$\pm$6.8} &  44.8 \\
        & Prompt-tuning &  
        51.9\footnotesize{$\pm$5.0} & 19.9\footnotesize{$\pm$2.4} & 51.6\footnotesize{$\pm$2.3} & 53.5\footnotesize{$\pm$2.1} & 52.7\footnotesize{$\pm$3.7} & 22.3\footnotesize{$\pm$9.0} & 34.4\footnotesize{$\pm$1.8} & 50.2\footnotesize{$\pm$1.5} & \textcolor{tblue}{\textbf{49.0}}\footnotesize{$\pm$12.9} & 42.8 \\
        & STT (ours) & \textcolor{tblue}{\textbf{61.9}}\footnotesize{$\pm$2.7} & \textcolor{tblue}{\textbf{26.0}}\footnotesize{$\pm$3.5} & \textcolor{tblue}{\textbf{53.9}}\footnotesize{$\pm$4.4} & \textcolor{tblue}{\textbf{60.3}}\footnotesize{$\pm$4.2} & \textcolor{tblue}{\textbf{60.6}}\footnotesize{$\pm$3.3} & 23.6\footnotesize{$\pm$5.5} & 33.4\footnotesize{$\pm$1.1} & \textcolor{tblue}{\textbf{51.6}}\footnotesize{$\pm$0.9} & 48.1\footnotesize{$\pm$9.4} & \textcolor{tblue}{\textbf{46.6}}\\
    \midrule[0.8pt]
        \multirow{5}{*}{\rotatebox[origin=c]{90}{\textbf{Five-shot}}}
        & Fine-tuning & 
        60.0\footnotesize{$\pm$5.1} & 32.3\footnotesize{$\pm$2.1} & 58.2\footnotesize{$\pm$3.7} & 63.9\footnotesize{$\pm$6.0} & 59.2\footnotesize{$\pm$1.5} & 61.3\footnotesize{$\pm$12.2} & 36.8\footnotesize{$\pm$2.6} & 53.9\footnotesize{$\pm$3.3} & 53.5\footnotesize{$\pm$5.2} & 53.2\\
        \cmidrule(lr){2-12}
        & Adapter & 49.8 \footnotesize{$\pm$0.9} & 17.9\footnotesize{$\pm$6.5} & 50.0\footnotesize{$\pm$0.0} & 50.0\footnotesize{$\pm$0.0} & 50.0\footnotesize{$\pm$0.1} & 16.2\footnotesize{$\pm$9.8} & 33.2\footnotesize{$\pm$0.4} & 49.5\footnotesize{$\pm$0.1} & 40.5\footnotesize{$\pm$11.2} & 40.5\\
        \cmidrule(lr){2-12}
        & Prefix-tuning & 
        54.9\footnotesize{$\pm$5.8} & 25.4\footnotesize{$\pm$1.3} & 54.3\footnotesize{$\pm$3.0} & 56.4\footnotesize{$\pm$4.0} & 56.2\footnotesize{$\pm$3.7} & \textcolor{tblue}{\textbf{29.6}}\footnotesize{$\pm$14.0} & 34.4\footnotesize{$\pm$1.1} & 51.5\footnotesize{$\pm$1.0} & \textcolor{tblue}{\textbf{46.0}}\footnotesize{$\pm$10.3} & 45.4 \\
        & Prompt-tuning &  
        52.2\footnotesize{$\pm$6.7} & 20.6\footnotesize{$\pm$3.5} & 55.6\footnotesize{$\pm$6.4} & 56.4\footnotesize{$\pm$4.4} & 53.7\footnotesize{$\pm$2.2} & 21.2\footnotesize{$\pm$5.7} & \textcolor{tblue}{\textbf{35.8}}\footnotesize{$\pm$1.6} & 51.8\footnotesize{$\pm$1.7} & 47.8\footnotesize{$\pm$9.5} & 43.9\\
        & STT (ours) & \textcolor{tblue}{\textbf{67.7}}\footnotesize{$\pm$5.1} & \textcolor{tblue}{\textbf{30.6}}\footnotesize{$\pm$2.1} & \textcolor{tblue}{\textbf{62.1}}\footnotesize{$\pm$3.0} & \textcolor{tblue}{\textbf{62.9}}\footnotesize{$\pm$4.1} & \textcolor{tblue}{\textbf{61.5}}\footnotesize{$\pm$2.3} & 25.4\footnotesize{$\pm$3.9} & 33.1\footnotesize{$\pm$0.6} & \textcolor{tblue}{\textbf{51.9}}\footnotesize{$\pm$0.7} & 44.3\footnotesize{$\pm$12.9} & \textcolor{tblue}{\textbf{48.9}}\\
    \bottomrule[1.2pt]
    \end{tabular}}
    \vspace{-2mm}
    \caption{Evaluation Results (mean and variance over 5 random trails) for one-shot, two-shot, and five-shot settings. The best results across the prompt-based methods (Prefix-tuning, Prompt-tuning, and STT) are shown in bold. Our STT approach exceeds both Prefix-tuning and Prompt-tuning in most tasks and on average.}
    \vspace{-2mm}
    \label{tab:results}
\end{table*}

\section{Soft Template Tuning}

Our task is to adapt a pre-trained masked language model to downstream tasks with a dataset $\mathcal{D}$. To realize the few-shot regime, we only sample $K$ examples from $\mathcal{D}$ as the training dataset $\mathcal{D}_{\text{train}} = {(x_i, y_i)}_{i=1}^{K}$, and use the original test dataset $\mathcal{D}_{\text{test}}$ for evaluation. We define $F: \mathcal{Y} \longrightarrow \mathcal{V}$ as a mapping from the label space $\mathcal{Y}$ to the word space with a vocabulary $\mathcal{V}$ in the pre-trained language model.

\paragraph{Manual Template Design}
We first construct prompted input $\hat{x}_{i}$ by adding manual prompts to the input data $x_i$. Inspired by \citet{gao2020making}, we adapt manual prompts for different tasks, which are defined in Appendix Table \ref{tab:manual_templates}. As an example, we augment an input $x_1$ ({\it e.g.}, \textit{The snacks are delicious.}) as follows for the classification task:
\begin{equation*}
    \hat{x}_{1} = \text{[CLS] $x_1$ it was [MASK] . [SEP]}
\end{equation*}
\vspace{-1mm}
After obtaining $\hat{x}_{1}$, we translate it into a hidden vector representation with the pre-trained embedding layer of the language model. 

\paragraph{Soft Prompt Tuning}
We next combine the manual template with a soft prompt to enrich the input. 
We sample $M$ words $\bm{Z} \triangleq (z_1, \cdots, z_M)$ from the vocabulary of the pre-trained language model, and then initialize a random embedding layer to represent the input $\bm{Z}$ as hidden vectors, which are learned during the tuning process. 

We then concatenate the hidden representations of the manual prompts and learnable prompts to form a new hidden vector, which is fed as the input to the transformer layers of language model. Instead of adopting the classification objective in the standard prompt-based methods, we close the gap between pre-training and fine-tuning by treating the tuning task as an masked language model (MLM) task. The probability of predicting the corresponding class $y$ is defined as
\begin{equation}
\begin{aligned}
p(y \mid x) &= p(\text{[MASK]} = F(y) \mid \hat{x},\bm{Z}) \\
&= \frac{\exp (\bm{w}_{F(y)} \cdot \bm{h}_{\text{[MASK]}})}{\sum_{y' \in \mathcal{Y}}{\exp (\bm{w}_{F(y')} \cdot \bm{h}_{\text{[MASK]}})}},
\end{aligned}
\end{equation}
where $\bm{h}_{\text{[MASK]}}$ is the hidden vector of [MASK] and $\bm{w}_{F(y)}$ denotes the linear weights before softmax function for class $y$. It is important to note that we have re-used the language model linear weights from the pre-trained language model. We also show in Section \ref{sec:app_lm_head} Ablation Study that tuning the language model head could gain significant improvement.

\begin{savenotes}
\begin{table}[t]
    \centering
    \scalebox{0.7}{
    \begin{tabular}{c|c c c c}
    \toprule
         \textbf{Models} & \makecell{\textbf{Embedding}\\\textbf{Layers}} & \makecell{\textbf{Transformer}\\\textbf{Layers}} & \makecell{\textbf{Head}\\\textbf{Layers}\footnote{This column is related to the number of task labels, and is calculated based on bi-classification.}} & \textbf{Total}\\
    \midrule
         Prefix-tuning & 0.026M & 20.752M &1.052M & \textbf{21.83M} \\
         Prompt-tuning & 0.026M & 0M & 1.052M & \textbf{1.08M}\\
         STT (ours) & 0.026M & 0M & 1.054M & \textbf{1.08M}\\
    \bottomrule
    \end{tabular}}
    \vspace{-2mm}
    \caption{Trainable parameters for the prompt-based methods (with prompt length 25). Note that, the total trainable parameters for fine-tuning is \textbf{355.36M}, which is much larger than all of the prompt-based methods.}
    \vspace{-4mm}
    \label{tab:parameters}
\end{table} 
\end{savenotes}

\vspace{-1mm}
\section{Experiments}
\vspace{-1mm}
\subsection{Experimental Setup}

\paragraph{Baselines}
We compared our proposed STT with the two recent prompt-based methods: prefix-tuning \citep{stanford} and prompt-tuning \citep{google}. We show structure differences between our model and these two prompt-based methods in Figure \ref{fig:model_structure}.
To illustrate the gap between prompt tuning and fine-tuning more clearly, we also provide the fine-tuning results. All of the baselines in our paper are based on the RoBERTa-large \citep{roberta} model in the HuggingFace Transformers codebase \citep{wolf-etal-2020-transformers}.

\vspace{-2mm}
\paragraph{Evaluation Settings}
Without loss of fairness, we evaluate our STT approach based on the pre-trained RoBERTa-large \citep{roberta} model as per the baseline methods. To realize the few-shot setting, for each task, evaluation is conducted by training on $K$ samples per class ($K=1,2,5$). We tested all the models with 5 seeds (13, 21, 42, 87, 100) and calculated mean and variance.

\paragraph{Hyper-parameters}
Hyper-parameters tuning is performed with a small development set, which is set to the same size as the training set. As pointed out by \citet{gao2020making}, this can avoid the significant advantages of using large development set and complies better with the goal of few-shot setting. For simplicity, instead of tuning hyper-parameters for each task, we only tune and select the hyper-parameters on SST-2, and apply the tuned parameters to remaining tasks. For each task, we conduct training for 500 steps, with a batch size of 2, a learning rate of $2e^{-5}$, and a prompt length of 25. 
Furthermore, the ameliorate unstable performance induced by the randomness of sampling a small dataset, we repeat each task and average the performance with variance over 5 random trials.

\begin{table*}[ht!]
    \centering
    \scalebox{0.85}{
    \begin{tabular}{l l l l}
    \toprule
    \textbf{Task} & \textbf{Template} & \textbf{Label Words} & \textbf{Domain}\\
    \midrule
    \multicolumn{4}{l}{\textbf{Sentiment Analysis Tasks}} \\
       SST-2 &  <S1> it was [MASK] . &positive: great, negative: terrible & movie reviews\\
       SST-5 & <S1> it was [MASK] . & positive: great, neutral: okay, negative: terrible & movie reviews\\
       MR & <S1> it was [MASK] . & positive: great, neutral: okay, negative: terrible & movie reviews\\
       CR & <S1> it was [MASK] . & positive: great, neutral: okay, negative: terrible & customer reviews\\
       MPQA & <S1> it was [MASK] . &positive: great, negative: terrible & news opinions\\
    \midrule
    \multicolumn{4}{l}{\textbf{Question Classification Tasks}} \\
       TREC & [MASK] : <S1> & \makecell[l]{abbreviation: Expression, entity: Entity,\\ description: Description, human: Human,\\ location: Location, numeric: Number} & misc.\\
    \midrule
    \multicolumn{4}{l}{\textbf{Inference Tasks}} \\
       SNLI & <S1> ? [MASK] , <S2> & entailment: yes, neutral: maybe, contradiction: no & misc.\\
       QNLI & <S1> ? [MASK] , <S2> & entailment: yes, not\_entailment: no & Wikipedia\\
    \midrule
    \multicolumn{4}{l}{\textbf{Similarity Tasks}} \\
       QQP & <S1> [MASK] , <S2> & equivalent: yes, not\_equivalent: no & social QA questions\\
    \bottomrule
    \end{tabular}}
    \caption{Task types and domains in our experiments with manual templates and label words.}
    \label{tab:manual_templates}
\end{table*}

\vspace{-2mm}
\subsection{Evaluation Tasks}

To conduct a systematic evaluation, we consider a total of 9 tasks, covering different domains and difficulties. Broadly speaking, the evaluation set includes 6 single-sentence tasks and 3 sentence-pair tasks. 


Specifically, the single-sentence tasks, including SST-2, SST-5 \citep{socher2013recursive}, MR \citep{pang2005seeing}, CR \citep{hu2004mining}, MPQA \citep{wiebe2005annotating}, and TREC \citep{voorhees2000building}, are used to test the model's performance on predicting the label word for each input sentence. The sentence-pair tasks, including SNLI \citep{bowman2015large}, QNLI \citep{rajpurkar2016squad}, QQP \citep{iyer2017first}, are introduced to measure how well a model is by comparing the relationships between 2 input sentences. Among them, SST-2, QQP, and QNLI are each selected from one category of the GLUE benchmark \citep{wang2018glue}. 

As summarized in Table~\ref{tab:manual_templates}, we select the evaluation tasks of various types and cover diverse domains. SST-2, SST-5, MR, CR, and MPQA belongs to sentiment analysis. TREC classifies open-domain, fact-based questions into different classes. SNLI and QNLI, respectively, are datasets for the inference of sentence relations and question answers. And QQP is a sentence similarity task from the social question-and-answer community Quora. In formulating them as masked language modeling task, we utilize manual templates and label words intuitively designed for each task (Table~\ref{tab:manual_templates}).

We follow \citet{gao2020making} for pre-processing and use the same way to sample the testing set.


\section{Results}

\vspace{-2mm}
\subsection{Main Results}
\vspace{-2mm}
Table \ref{tab:results} (Top) shows the one-shot evaluation results. Regarding the number of trainable parameters, our model only tunes 1.08M parameters, which is almost the same as prompt-tuning \cite{google}, less than prefix-tuning \cite{stanford}, and far less than fine-tuning.
In this set of experiments, although prefix-tuning and prompt-tuning surpass fine-tuning in some tasks, they still lag behind fine-tuning on average. Comparing with the prompt-tuning method \citep{google}, prefix-tuning gains 0.6\% performance improvement, but at the expense of introducing additional prefix parameters to each transformer layer. 
Our method, on the basis of prompt-tuning \cite{google}, improves the average accuracy by 2.2 points. With only 0.3\% trainable parameters, it even surpasses the fine-tuning method in most tasks, resulting in a 0.7 points improvement on average. 

Table \ref{tab:results} (Middle and Bottom) shows the evaluation results for two-shot and five-shot scenarios. Similarly, our STT approach surpasses prefix-tuning \cite{stanford} and prompt-tuning \cite{google} in general. Surprisingly, when compared with fine-tuning, our method still achieves significantly better results on SST-2, MR, and MPQA.

Table \ref{tab:parameters} provides a comparison on model parameters. Although language model head uses a linear transformation for mapping hidden vector to dictionary size, we only utilize the matrix with indices of the label words (which will be 2 for bi-classification task, 3 for tri-classification task, 6 for TREC task).

Discussion could be found in Section \ref{sec:discussion}.

\vspace{-1mm}
\subsection{Ablation Studies}

\begin{figure}[t]
    \centering
    \begin{subfigure}{.25\textwidth}
        \centering
        \includegraphics[width=0.95\textwidth]{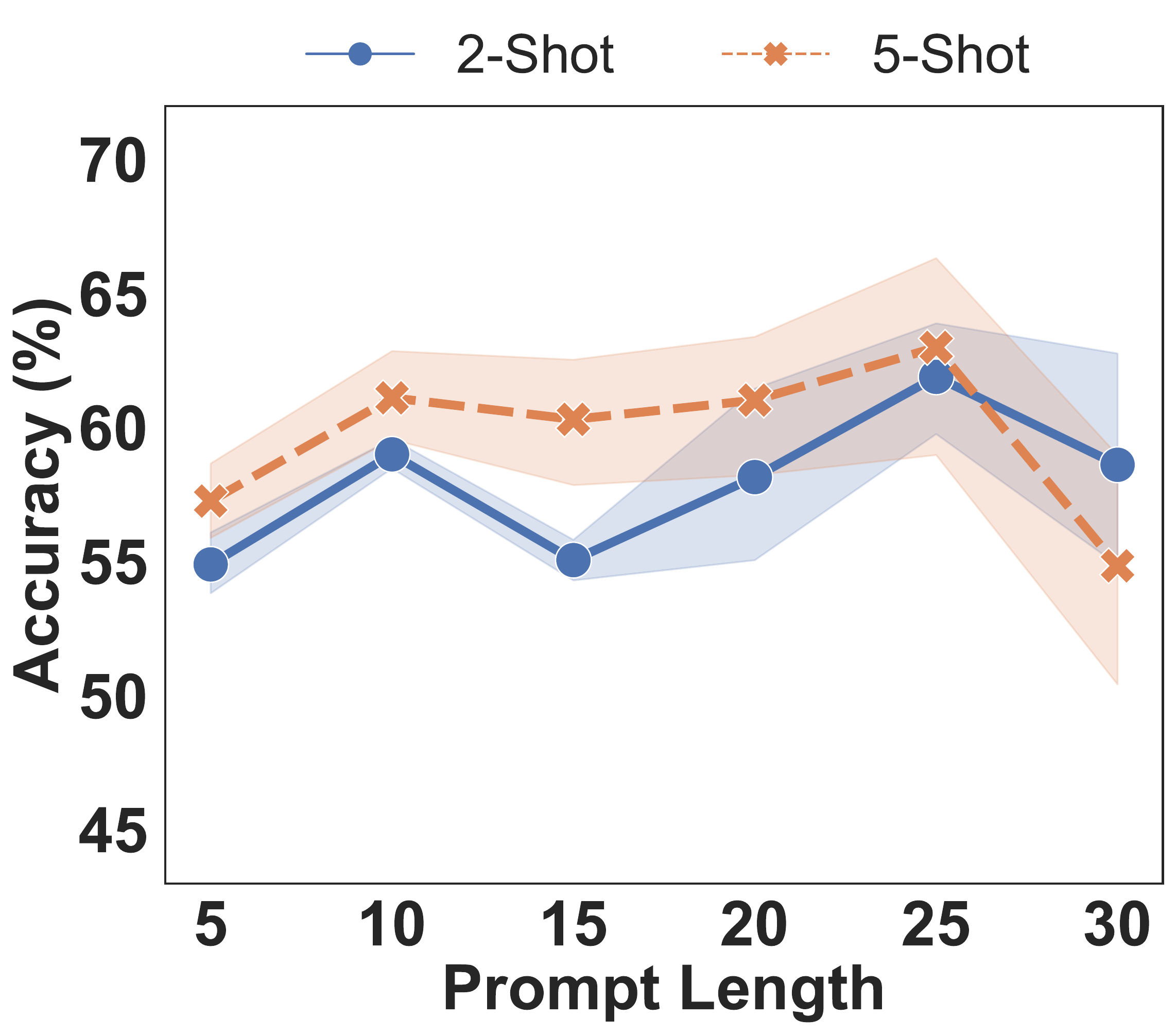}
    \end{subfigure}%
    \begin{subfigure}{.25\textwidth}
        \centering
        \includegraphics[width=0.95\textwidth]{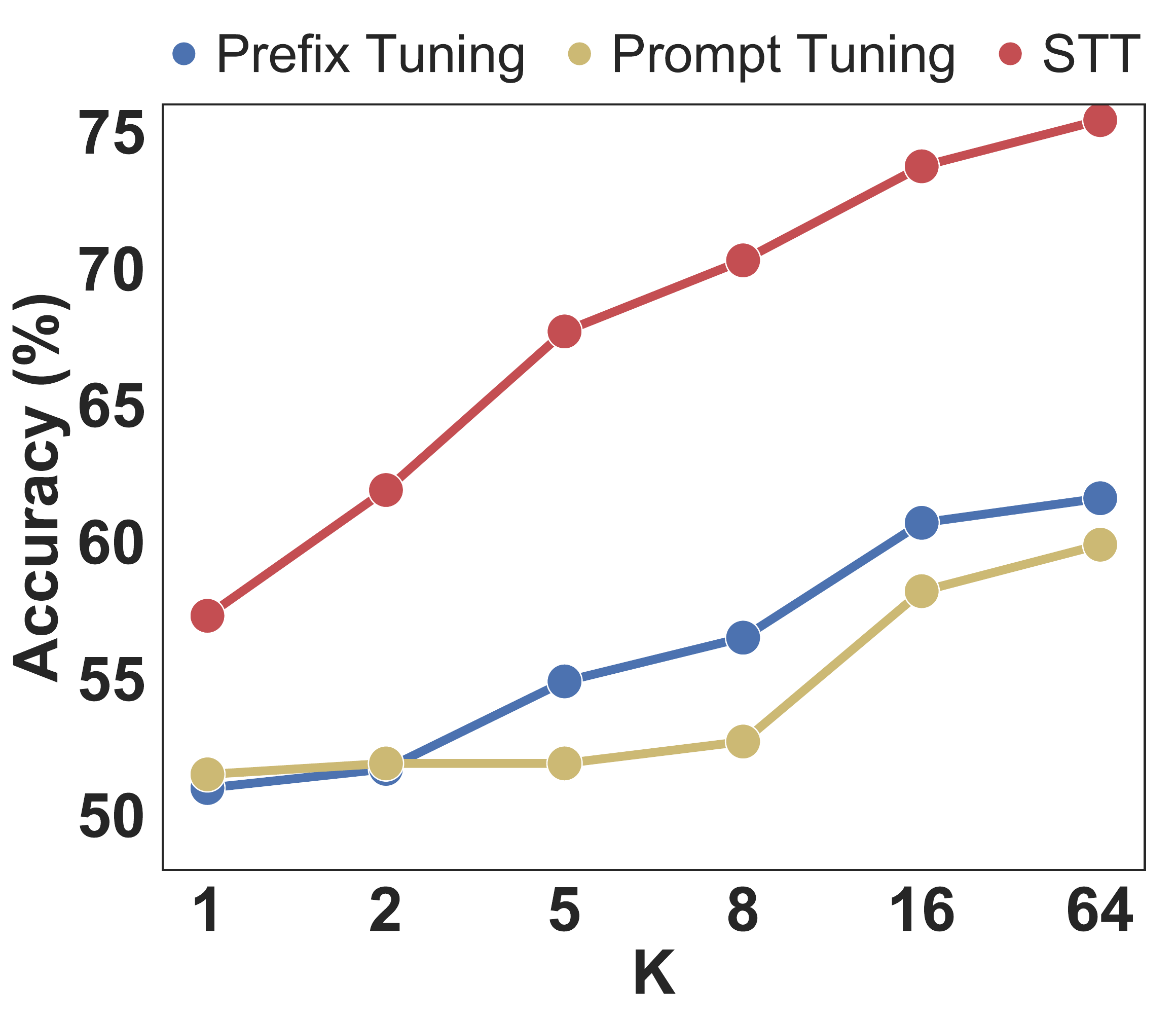}
    \end{subfigure}
    \vspace{-2mm}
    \caption{\textbf{Left:} Performance on SST-2 task over different prompt lengths. A prompt length of 25 generally gives better results. \textbf{Right:} Performance comparison over different $K$ (\# samples per class) size.}
    \vspace{-3mm}
    \label{fig:ablation}
\end{figure}

\paragraph{Prompt Length}
Prompt length is a critical hyper-parameter for prompt-based methods. We therefore test the impact of prompt length on our STT. We initialize STT with various prompt lengths from 5 to 30, with an increment of 5. Intuitively, more prompt tokens indicate better expressive power but introduce slightly more trainable parameters. The results are plotted in Figure \ref{fig:ablation} (left). We observe that the best results are achieved with around a certain prompt length (in our case, 25 for both 2-shot and 5-shot settings), while inserting more prompt tokens beyond this point may yield a performance drop, especially in the few-shot setting.

\paragraph{\textbf{$K$} Size} 
We also evaluate how well STT works by increasing $K$ in comparison with Prefix Tuning \citep{stanford} and Prompt Tuning \citep{google}. As shown in Figure \ref{fig:ablation} (Right), with $K$ increasing all the way up to 64, the performance of STT keeps increasing, which consistently outperforms the other prompt-based methods, demonstrating a great potential when more training data become available.

\paragraph{The Role of Trainable Language Model Head}
\label{sec:app_lm_head}
To demonstrate the importance of tuning the language model head, we conducted a comparative experiment between making the language model head fixed or trainable for tuning. The experiment was performed on SST-2 task over five random trials, and with a prompt length of 25. The results are concluded in Table~\ref{tab:update_lm_head}. Tuning with language model head updated makes a marked improvement in all experiments, illustrating that it is essential to make the language model head trainable for our method.

\begin{table}[t]
    \centering
    \scalebox{0.6}{
    \begin{tabular}{c|c c c c c|c}
    \toprule
         \textbf{Seeds} & \textbf{13} & \textbf{21} & \textbf{42} & \textbf{87} & \textbf{100} & \textbf{Average}\\
    \midrule
         w/o updated \textit{lm\_head} & 53.56 & 53.78 & 53.78 & 53.33 & 53.55 & 53.56{$\pm$0.2} \\
         w/ updated \textit{lm\_head} & \textcolor{tblue}{\textbf{62.73}} & \textcolor{tblue}{\textbf{63.07}} & \textcolor{tblue}{\textbf{68.92}} & \textcolor{tblue}{\textbf{55.96}} & \textcolor{tblue}{\textbf{64.33}} & \textcolor{tblue}{\textbf{63.00}}{$\pm$4.6}\\
    \bottomrule
    \end{tabular}}
    \vspace{-2mm}
    \caption{Comparison between tuning with trainable \textit{lm\_head} and fixed \textit{lm\_head}. \textit{lm\_head} stands for language model head.}
    \vspace{-4mm}
    \label{tab:update_lm_head}
\end{table}

\section{Discussion}
\label{sec:discussion}
We find that existing prompt-based methods do not work well in this setting, whereas our STT can dramatically improve over the prompt-based methods without introducing additional parameters. Although our trainable parameters are much less than fine-tuning, STT can still outperform the fine-tuning method on SST-2, MR, MPQA. Most of these tasks belong to sentiment classification tasks. The performance improvement over the fine-tuning method may be attributed to the benefit of the manually-designed templates. Therefore, the performance of STT depends on the quality of manual templates and label words choice, which can be one limitation of our method. 

In our experiments, we choose a public pre-trained model (RoBERTa large) in the huggingface transformer library with 355.36M parameters in total, which is a moderate-sized pre-trained language model. As the model size increases, the advantage of the prompt-based models over fine-tuning will be more obvious. \citet{google} concludes that the gap between fine-tuning and prompt-based methods is closing when increasing the pre-trained language model size. Prompt-tuning method \citep{google} could achieve competitive results when the model size reaches 100B. We will validate our STT on a larger pre-trained language model for future work.

For ethical considerations, we need to point out, that bias might be introduced by training. This, however, is owing to the limitations of the dataset 
\footnote{As pointed out in \url{https://huggingface.co/roberta-large}, the training data contains unfiltered and un-neutral content from the internet. Therefore, the model can have biased predictions.} 
and few-shot training samples at hand, and does not represent a flaw in the model.

\subsection{Related Works}
\label{sec:related_works}



\paragraph{Language Model Fine-tuning}
A prevalent idea of adapting pre-trained language models for a wide range of downstream tasks is to fine-tune the pre-trained models with task specific heads. Since it became a standard practise of transfer learning in NLP domain, a number of efforts have been made seeking better ways of language model fine-tuning \citep{howard2018universal, gururangan2020don, phang2018sentence, aghajanyan2021muppet, dodge2020fine, pruksachatkun2020intermediate, zhang2020revisiting}. However, fine-tuning usually requires the updating of all the model parameters, and therefore, a copy of the whole model needs to be stored for each specific downstream task. It is hence expensive for both computational and storage resources. Another significant problem lies in the different objective formats between the pre-training stage and fine-tuning stage. This not only leaves a gap between the language model and downstream tasks, but also introduces new parameters, making the fine-tuning less effective in the few-shot setting.

\paragraph{Prompting}
Prompting was recently proposed aiming at the above issues, especially for giant language models like GPT-3 \citep{gpt3}. A prompt is usually referred to as some extra text information added to the input. By concatenating the input with a sequence of tokens on which the language model could condition, prompting enables the downstream tasks to adopt the same type of objective as the pre-training stage, and therefore, closes the gap between the 2 stages. Taking sentiment analysis as an example, the input sentence is concatenated with a prompt (e.g., ``it was [MASK]''), where the label word is masked and to be predicted by the language model. Then the sentiment class could be inferred based on which of the selected label word is predicted (e.g., ``great'' as positive and ``terrible'' as negative). In addition, since prompting introduces no new parameters and requires no training (all parameters are fixed), it was demonstrated to be effective in the few-shot setting \citep{le2021many}, and designing or searching for optimal prompts becomes a crucial issue \citep{schick2020exploiting, shin2020autoprompt}.

\paragraph{Prompt-tuning}
Instead of adding discrete prompt tokens from the vocabulary, prompt-tuning uses soft prompts (in the form of trainable continuous embeddings) and achieved significant improvements over prompting in many tasks \citep{google}. Specifically, \citet{stanford} proposed prefix-tuning for conditional generation tasks, where continuous embeddings were prepended to each layer of the encoder (and decoder if applicable) architecture as prefix. By tuning only the prefix parameters, comparable performance to fine-tuning was observed in full dataset setting. Some other tasks \citep{hambardzumyan2021warp, google} further simplified prefix-tuning by excluding intermediate-layer prefixes and only restricting the trainable parameters to the input. Soft prompts were also proved to be effective in knowledge probing tasks when inserting prompts into different positions of the input according to a manually determined pattern \citep{qin2021learning, liu2021gpt}. 
\textbf{P-tuning} \cite{liu2021gpt} demonstrated that by searching soft prompts in the continuous space, GPT-style models could achieve competitive performance with similar-size BERTs in NLU tasks. It also claims that adding task-related anchor tokens could bring further improvement. However, P-tuning introduces extra parameters by using bidirectional LSTM for prompt embedding, and hence fine-tunes the prompt embeddings together with the the pre-trained model for SuperGLUE \cite{wang2019superglue} tasks. P-tuning v2 \cite{liu2021p} further enhanced the universality of P-tuning on a broader selection of models and tasks at the cost of introducing soft prompts (more trainable parameters) to every layer of the pre-trained model, and is therefore only evaluated in the fully-supervised setting rather than few-shot setting.
Despite prompt-tuning has demonstrated strength even comparable to fine-tuning when given enough data, we observed a considerable gap in performance between prompt-tuning and fine-tuning in the few-shot setting.
\citet{gu2021ppt} proposed Pre-trained Prompt Tuning framework (PPT) and proved that their approach could reach or even outperform full-model fine-tuning under few-shot settings. However, according to the format of task, they require that soft prompts need to be pre-trained on a designed self-supervised task before fine-tuning the prompts on the downstream task.

\vspace{-1mm}
\section{Conclusion}
\vspace{-1mm}
To adapt prompt-based methods in the few-shot regime, we propose STT, a simple method that combines both manual templates and soft prompt, and treats the downstream classification tasks as a language modeling task. We conduct experiments on 9 downstream classification tasks. Experiment results reveal that STT outperforms both fine-tuning and prompt-based methods under the one-shot setting. In the two-shot and five-shot settings, our approach closes the gap between the fine-tuning method and prompted-based methods, which can even outperform fine-tuning on classification tasks. Our research indicates that prompt-tuning is an effective tool for adapting large pre-trained models, yet still there is room for further improvement, especially in the few-shot regime.


\bibliography{anthology,custom}

\begin{thebibliography}{37}
\expandafter\ifx\csname natexlab\endcsname\relax\def\natexlab#1{#1}\fi

\bibitem[{Aghajanyan et~al.(2021)Aghajanyan, Gupta, Shrivastava, Chen,
  Zettlemoyer, and Gupta}]{aghajanyan2021muppet}
Armen Aghajanyan, Anchit Gupta, Akshat Shrivastava, Xilun Chen, Luke
  Zettlemoyer, and Sonal Gupta. 2021.
\newblock Muppet: Massive multi-task representations with pre-finetuning.
\newblock \emph{arXiv preprint arXiv:2101.11038}.

\bibitem[{Bowman et~al.(2015)Bowman, Angeli, Potts, and
  Manning}]{bowman2015large}
Samuel~R Bowman, Gabor Angeli, Christopher Potts, and Christopher~D Manning.
  2015.
\newblock A large annotated corpus for learning natural language inference.
\newblock \emph{arXiv preprint arXiv:1508.05326}.

\bibitem[{Brown et~al.(2020)Brown, Mann, Ryder, Subbiah, Kaplan, Dhariwal,
  Neelakantan, Shyam, Sastry, Askell et~al.}]{gpt3}
Tom~B Brown, Benjamin Mann, Nick Ryder, Melanie Subbiah, Jared Kaplan, Prafulla
  Dhariwal, Arvind Neelakantan, Pranav Shyam, Girish Sastry, Amanda Askell,
  et~al. 2020.
\newblock Language models are few-shot learners.
\newblock \emph{arXiv preprint arXiv:2005.14165}.

\bibitem[{Devlin et~al.(2018)Devlin, Chang, Lee, and
  Toutanova}]{devlin2018bert}
Jacob Devlin, Ming-Wei Chang, Kenton Lee, and Kristina Toutanova. 2018.
\newblock Bert: Pre-training of deep bidirectional transformers for language
  understanding.
\newblock \emph{arXiv preprint arXiv:1810.04805}.

\bibitem[{Dodge et~al.(2020)Dodge, Ilharco, Schwartz, Farhadi, Hajishirzi, and
  Smith}]{dodge2020fine}
Jesse Dodge, Gabriel Ilharco, Roy Schwartz, Ali Farhadi, Hannaneh Hajishirzi,
  and Noah Smith. 2020.
\newblock Fine-tuning pretrained language models: Weight initializations, data
  orders, and early stopping.
\newblock \emph{arXiv preprint arXiv:2002.06305}.

\bibitem[{Gao et~al.(2020)Gao, Fisch, and Chen}]{gao2020making}
Tianyu Gao, Adam Fisch, and Danqi Chen. 2020.
\newblock Making pre-trained language models better few-shot learners.
\newblock \emph{arXiv preprint arXiv:2012.15723}.

\bibitem[{Gu et~al.(2021)Gu, Han, Liu, and Huang}]{gu2021ppt}
Yuxian Gu, Xu~Han, Zhiyuan Liu, and Minlie Huang. 2021.
\newblock Ppt: Pre-trained prompt tuning for few-shot learning.
\newblock \emph{arXiv preprint arXiv:2109.04332}.

\bibitem[{Gururangan et~al.(2020)Gururangan, Marasovi{\'c}, Swayamdipta, Lo,
  Beltagy, Downey, and Smith}]{gururangan2020don}
Suchin Gururangan, Ana Marasovi{\'c}, Swabha Swayamdipta, Kyle Lo, Iz~Beltagy,
  Doug Downey, and Noah~A Smith. 2020.
\newblock Don't stop pretraining: adapt language models to domains and tasks.
\newblock \emph{arXiv preprint arXiv:2004.10964}.

\bibitem[{Hambardzumyan et~al.(2021)Hambardzumyan, Khachatrian, and
  May}]{hambardzumyan2021warp}
Karen Hambardzumyan, Hrant Khachatrian, and Jonathan May. 2021.
\newblock Warp: Word-level adversarial reprogramming.
\newblock \emph{arXiv preprint arXiv:2101.00121}.

\bibitem[{Han et~al.(2021)Han, Zhao, Ding, Liu, and Sun}]{han2021ptr}
Xu~Han, Weilin Zhao, Ning Ding, Zhiyuan Liu, and Maosong Sun. 2021.
\newblock Ptr: Prompt tuning with rules for text classification.
\newblock \emph{arXiv preprint arXiv:2105.11259}.

\bibitem[{Howard and Ruder(2018)}]{howard2018universal}
Jeremy Howard and Sebastian Ruder. 2018.
\newblock Universal language model fine-tuning for text classification.
\newblock \emph{arXiv preprint arXiv:1801.06146}.

\bibitem[{Hu and Liu(2004)}]{hu2004mining}
Minqing Hu and Bing Liu. 2004.
\newblock Mining and summarizing customer reviews.
\newblock In \emph{Proceedings of the tenth ACM SIGKDD international conference
  on Knowledge discovery and data mining}, pages 168--177.

\bibitem[{Iyer et~al.(2017)Iyer, Dandekar, Csernai et~al.}]{iyer2017first}
Shankar Iyer, Nikhil Dandekar, Korn{\'e}l Csernai, et~al. 2017.
\newblock First quora dataset release: Question pairs.
\newblock \emph{data. quora. com}.

\bibitem[{Le~Scao and Rush(2021)}]{le2021many}
Teven Le~Scao and Alexander~M Rush. 2021.
\newblock How many data points is a prompt worth?
\newblock In \emph{Proceedings of the 2021 Conference of the North American
  Chapter of the Association for Computational Linguistics: Human Language
  Technologies}, pages 2627--2636.

\bibitem[{Lester et~al.(2021)Lester, Al-Rfou, and Constant}]{google}
Brian Lester, Rami Al-Rfou, and Noah Constant. 2021.
\newblock The power of scale for parameter-efficient prompt tuning.
\newblock \emph{arXiv preprint arXiv:2104.08691}.

\bibitem[{Lewis et~al.(2019)Lewis, Liu, Goyal, Ghazvininejad, Mohamed, Levy,
  Stoyanov, and Zettlemoyer}]{lewis2019bart}
Mike Lewis, Yinhan Liu, Naman Goyal, Marjan Ghazvininejad, Abdelrahman Mohamed,
  Omer Levy, Ves Stoyanov, and Luke Zettlemoyer. 2019.
\newblock Bart: Denoising sequence-to-sequence pre-training for natural
  language generation, translation, and comprehension.
\newblock \emph{arXiv preprint arXiv:1910.13461}.

\bibitem[{Li and Liang(2021)}]{stanford}
Xiang~Lisa Li and Percy Liang. 2021.
\newblock Prefix-tuning: Optimizing continuous prompts for generation.
\newblock \emph{arXiv preprint arXiv:2101.00190}.

\bibitem[{Liu et~al.(2021{\natexlab{a}})Liu, Ji, Fu, Du, Yang, and
  Tang}]{liu2021p}
Xiao Liu, Kaixuan Ji, Yicheng Fu, Zhengxiao Du, Zhilin Yang, and Jie Tang.
  2021{\natexlab{a}}.
\newblock P-tuning v2: Prompt tuning can be comparable to fine-tuning
  universally across scales and tasks.
\newblock \emph{arXiv preprint arXiv:2110.07602}.

\bibitem[{Liu et~al.(2021{\natexlab{b}})Liu, Zheng, Du, Ding, Qian, Yang, and
  Tang}]{liu2021gpt}
Xiao Liu, Yanan Zheng, Zhengxiao Du, Ming Ding, Yujie Qian, Zhilin Yang, and
  Jie Tang. 2021{\natexlab{b}}.
\newblock Gpt understands, too.
\newblock \emph{arXiv preprint arXiv:2103.10385}.

\bibitem[{Liu et~al.(2019{\natexlab{a}})Liu, Ott, Goyal, Du, Joshi, Chen, Levy,
  Lewis, Zettlemoyer, and Stoyanov}]{liu2019roberta}
Yinhan Liu, Myle Ott, Naman Goyal, Jingfei Du, Mandar Joshi, Danqi Chen, Omer
  Levy, Mike Lewis, Luke Zettlemoyer, and Veselin Stoyanov. 2019{\natexlab{a}}.
\newblock Roberta: A robustly optimized bert pretraining approach.
\newblock \emph{arXiv preprint arXiv:1907.11692}.

\bibitem[{Liu et~al.(2019{\natexlab{b}})Liu, Ott, Goyal, Du, Joshi, Chen, Levy,
  Lewis, Zettlemoyer, and Stoyanov}]{roberta}
Yinhan Liu, Myle Ott, Naman Goyal, Jingfei Du, Mandar Joshi, Danqi Chen, Omer
  Levy, Mike Lewis, Luke Zettlemoyer, and Veselin Stoyanov. 2019{\natexlab{b}}.
\newblock Roberta: A robustly optimized bert pretraining approach.
\newblock \emph{arXiv preprint arXiv:1907.11692}.

\bibitem[{Pang and Lee(2005)}]{pang2005seeing}
Bo~Pang and Lillian Lee. 2005.
\newblock Seeing stars: Exploiting class relationships for sentiment
  categorization with respect to rating scales.
\newblock \emph{arXiv preprint cs/0506075}.

\bibitem[{Phang et~al.(2018)Phang, F{\'e}vry, and Bowman}]{phang2018sentence}
Jason Phang, Thibault F{\'e}vry, and Samuel~R Bowman. 2018.
\newblock Sentence encoders on stilts: Supplementary training on intermediate
  labeled-data tasks.
\newblock \emph{arXiv preprint arXiv:1811.01088}.

\bibitem[{Pruksachatkun et~al.(2020)Pruksachatkun, Phang, Liu, Htut, Zhang,
  Pang, Vania, Kann, and Bowman}]{pruksachatkun2020intermediate}
Yada Pruksachatkun, Jason Phang, Haokun Liu, Phu~Mon Htut, Xiaoyi Zhang,
  Richard~Yuanzhe Pang, Clara Vania, Katharina Kann, and Samuel~R Bowman. 2020.
\newblock Intermediate-task transfer learning with pretrained models for
  natural language understanding: When and why does it work?
\newblock \emph{arXiv preprint arXiv:2005.00628}.

\bibitem[{Qin and Eisner(2021)}]{qin2021learning}
Guanghui Qin and Jason Eisner. 2021.
\newblock Learning how to ask: Querying lms with mixtures of soft prompts.
\newblock \emph{arXiv preprint arXiv:2104.06599}.

\bibitem[{Radford et~al.(2018)Radford, Narasimhan, Salimans, and
  Sutskever}]{radford2018improving}
Alec Radford, Karthik Narasimhan, Tim Salimans, and Ilya Sutskever. 2018.
\newblock Improving language understanding by generative pre-training.

\bibitem[{Raffel et~al.(2019)Raffel, Shazeer, Roberts, Lee, Narang, Matena,
  Zhou, Li, and Liu}]{raffel2019exploring}
Colin Raffel, Noam Shazeer, Adam Roberts, Katherine Lee, Sharan Narang, Michael
  Matena, Yanqi Zhou, Wei Li, and Peter~J Liu. 2019.
\newblock Exploring the limits of transfer learning with a unified text-to-text
  transformer.
\newblock \emph{arXiv preprint arXiv:1910.10683}.

\bibitem[{Rajpurkar et~al.(2016)Rajpurkar, Zhang, Lopyrev, and
  Liang}]{rajpurkar2016squad}
Pranav Rajpurkar, Jian Zhang, Konstantin Lopyrev, and Percy Liang. 2016.
\newblock Squad: 100,000+ questions for machine comprehension of text.
\newblock \emph{arXiv preprint arXiv:1606.05250}.

\bibitem[{Schick and Sch{\"u}tze(2020)}]{schick2020exploiting}
Timo Schick and Hinrich Sch{\"u}tze. 2020.
\newblock Exploiting cloze questions for few shot text classification and
  natural language inference.
\newblock \emph{arXiv preprint arXiv:2001.07676}.

\bibitem[{Shin et~al.(2020)Shin, Razeghi, Logan~IV, Wallace, and
  Singh}]{shin2020autoprompt}
Taylor Shin, Yasaman Razeghi, Robert~L Logan~IV, Eric Wallace, and Sameer
  Singh. 2020.
\newblock Autoprompt: Eliciting knowledge from language models with
  automatically generated prompts.
\newblock \emph{arXiv preprint arXiv:2010.15980}.

\bibitem[{Socher et~al.(2013)Socher, Perelygin, Wu, Chuang, Manning, Ng, and
  Potts}]{socher2013recursive}
Richard Socher, Alex Perelygin, Jean Wu, Jason Chuang, Christopher~D Manning,
  Andrew~Y Ng, and Christopher Potts. 2013.
\newblock Recursive deep models for semantic compositionality over a sentiment
  treebank.
\newblock In \emph{Proceedings of the 2013 conference on empirical methods in
  natural language processing}, pages 1631--1642.

\bibitem[{Voorhees and Tice(2000)}]{voorhees2000building}
Ellen~M Voorhees and Dawn~M Tice. 2000.
\newblock Building a question answering test collection.
\newblock In \emph{Proceedings of the 23rd annual international ACM SIGIR
  conference on Research and development in information retrieval}, pages
  200--207.

\bibitem[{Wang et~al.(2019)Wang, Pruksachatkun, Nangia, Singh, Michael, Hill,
  Levy, and Bowman}]{wang2019superglue}
Alex Wang, Yada Pruksachatkun, Nikita Nangia, Amanpreet Singh, Julian Michael,
  Felix Hill, Omer Levy, and Samuel Bowman. 2019.
\newblock Superglue: A stickier benchmark for general-purpose language
  understanding systems.
\newblock \emph{Advances in neural information processing systems}, 32.

\bibitem[{Wang et~al.(2018)Wang, Singh, Michael, Hill, Levy, and
  Bowman}]{wang2018glue}
Alex Wang, Amanpreet Singh, Julian Michael, Felix Hill, Omer Levy, and Samuel~R
  Bowman. 2018.
\newblock Glue: A multi-task benchmark and analysis platform for natural
  language understanding.
\newblock \emph{arXiv preprint arXiv:1804.07461}.

\bibitem[{Wiebe et~al.(2005)Wiebe, Wilson, and Cardie}]{wiebe2005annotating}
Janyce Wiebe, Theresa Wilson, and Claire Cardie. 2005.
\newblock Annotating expressions of opinions and emotions in language.
\newblock \emph{Language resources and evaluation}, 39(2):165--210.

\bibitem[{Wolf et~al.(2020)Wolf, Debut, Sanh, Chaumond, Delangue, Moi, Cistac,
  Rault, Louf, Funtowicz, Davison, Shleifer, von Platen, Ma, Jernite, Plu, Xu,
  Scao, Gugger, Drame, Lhoest, and Rush}]{wolf-etal-2020-transformers}
Thomas Wolf, Lysandre Debut, Victor Sanh, Julien Chaumond, Clement Delangue,
  Anthony Moi, Pierric Cistac, Tim Rault, Rémi Louf, Morgan Funtowicz, Joe
  Davison, Sam Shleifer, Patrick von Platen, Clara Ma, Yacine Jernite, Julien
  Plu, Canwen Xu, Teven~Le Scao, Sylvain Gugger, Mariama Drame, Quentin Lhoest,
  and Alexander~M. Rush. 2020.
\newblock \href {https://www.aclweb.org/anthology/2020.emnlp-demos.6}
  {Transformers: State-of-the-art natural language processing}.
\newblock In \emph{Proceedings of the 2020 Conference on Empirical Methods in
  Natural Language Processing: System Demonstrations}, pages 38--45, Online.
  Association for Computational Linguistics.

\bibitem[{Zhang et~al.(2020)Zhang, Wu, Katiyar, Weinberger, and
  Artzi}]{zhang2020revisiting}
Tianyi Zhang, Felix Wu, Arzoo Katiyar, Kilian~Q Weinberger, and Yoav Artzi.
  2020.
\newblock Revisiting few-sample bert fine-tuning.
\newblock \emph{arXiv preprint arXiv:2006.05987}.

\end{thebibliography}
\bibliographystyle{acl_natbib}




\end{document}